\documentclass[conference]{IEEEtran}
\IEEEoverridecommandlockouts
\usepackage[utf8]{inputenc} 
\usepackage[T1]{fontenc}    
\usepackage{hyperref}       
\usepackage{url}            
\usepackage{booktabs}       
\usepackage{amsfonts}       
\usepackage{nicefrac}       
\usepackage{microtype}      
\usepackage{xcolor}         
\usepackage{algorithm}
\usepackage{algpseudocode}
\usepackage{amsmath}
\usepackage{natbib}
\usepackage{tikz}
\usetikzlibrary{shapes.geometric, arrows, positioning}
\usepackage{graphicx}
\usepackage{wrapfig}
\usepackage{amssymb}
\usepackage{lipsum} 
\usepackage{xr}
\def\BibTeX{{\rm B\kern-.05em{\sc i\kern-.025em b}\kern-.08em
    T\kern-.1667em\lower.7ex\hbox{E}\kern-.125emX}}
\begin{document}

\title{SketcherX: AI-Driven Interactive Robotic drawing with Diffusion model and Vectorization Techniques\\}

\author{\IEEEauthorblockN{Jookyung Song}
\IEEEauthorblockA{\textit{Xorbis Co., Ltd.} \\
\textit{Seoul National University}\\
chsjk9005@snu.ac.kr}
\and
\IEEEauthorblockN{Mookyoung Kang}
\IEEEauthorblockA{\textit{Xorbis Co., Ltd.} \\
nostone@gmail.com}
\and
\IEEEauthorblockN{Nojun Kwak}
\IEEEauthorblockA{\textit{Seoul National University}\\
nojunk@snu.ac.kr}
}

\maketitle
\vspace{-1cm}
\begin{figure*}[ht!]
\centering
\makebox[\textwidth]{\includegraphics[width=\textwidth]{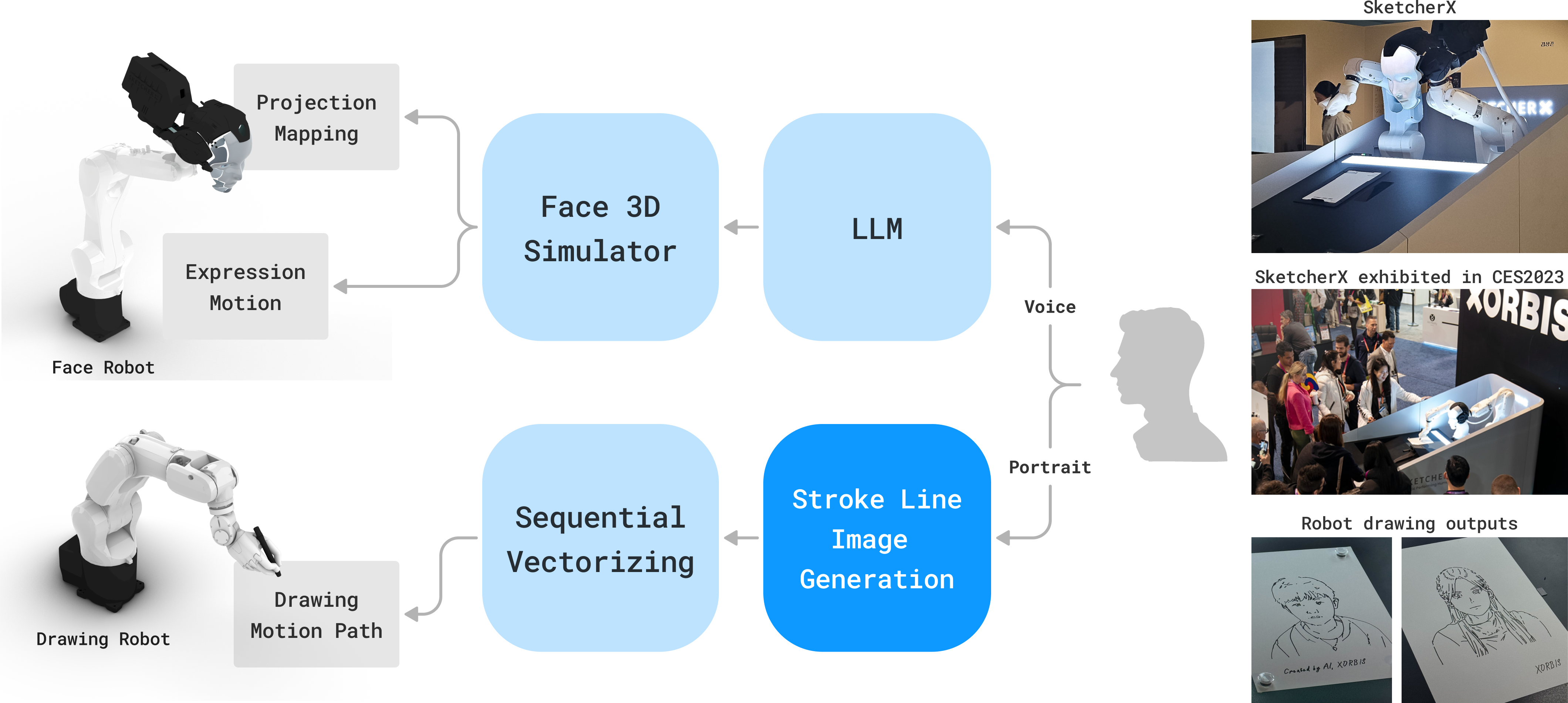}}
\caption{Overview of the SketcherX process. The top-right section shows an actual photograph of the SketcherX, CES 2023 exhibition, and examples of actual drawing. 
SketcherX consists of two robotic arms: one interacts with users using a camera and LLM, while the other processes images to create stylized, vectorized portraits.}
\label{fig1}
\end{figure*}

\begin{abstract}
We introduce SketcherX, a novel robotic system for personalized portrait drawing through interactive human-robot engagement. Unlike traditional robotic art systems that rely on analog printing techniques, SketcherX captures and processes facial images to produce vectorized drawings in a distinctive, human-like artistic style. The system comprises two 6-axis robotic arms : a face robot, which is equipped with a head-mounted camera and Large Language Model (LLM) for real-time interaction, and a drawing robot, utilizing a fine-tuned Stable Diffusion model, ControlNet, and Vision-Language models for dynamic, stylized drawing. Our contributions include the development of a custom Vector Low Rank Adaptation model (LoRA), enabling seamless adaptation to various artistic styles, and integrating a pair-wise fine-tuning approach to enhance stroke quality and stylistic accuracy. Experimental results demonstrate the system's ability to produce high-quality, personalized portraits within two minutes, highlighting its potential as a new paradigm in robotic creativity. This work advances the field of robotic art by positioning robots as active participants in the creative process, paving the way for future explorations in interactive, human-robot artistic collaboration.
\end{abstract}


\section{Introduction}
SketcherX is a robotic system designed to interact with people by drawing their portraits. When a user sits in front of the robot, a camera mounted on the robot's head captures the user's face. This captured image is then transformed into the robot's distinctive drawing style. The transformed image is further converted into vector strokes, enabling the robot's arm to draw the portrait like a human artist would. This interactive experience allows users to witness the creative act of the robot, redefining the value of robots as interactive media. 
As shown in Fig.~\ref{fig1}, SketcherX was successfully exhibited at CES 2023 and attracted a lot of attention. \footnote{'SketcherX at CES 2023', YouTube, uploaded by Xorbis LAB. Available at: \url{https://www.youtube.com/watch?v=4hr6Y67IQVA\&t=1s}}
 
While robotic drawing is not a new concept, previous attempts, such as those by \citep{birsak2018string} and \citep{robandnick2024darkfactory}, primarily involved robots recording bitmap data onto analog media using predefined rules, a process more akin to "printing" than true "drawing." Some research, such as Frida \citep{schaldenbrand2023frida}, has focused on imbuing robotic movements with creative significance and the ability to perform artistic acts. However, the outcomes of these efforts, while considered artworks, often lack a distinct robotic identity.

Our objective is to ensure that the drawings produced by SketcherX are perceived as creations of an autonomous entity. This requires the following characteristics in the media creations:

1) The creating entity must have an independent identity and interact with people.

2) The process of creation should be demonstrated in a manner comprehensible to humans.

Considering these characteristics, we developed two robotic arms: one with a face that can interact with people and the other capable of drawing. The arm responsible for the face uses a large language model (LLM) to communicate with users and is equipped with a rear-projection mask to display facial movements. The robot's face is dynamically rendered using a 3D simulator built in Unity, allowing it to express emotions through facial expressions and gestures based on the context of the conversation.

The drawing arm is designed to effectively demonstrate and communicate the drawing process to the user. After capturing the facial features of the person, the image is processed through a diffusion-based image processing technique to transform it into the robot's unique drawing style. This raster image is then mapped into a vector image which is again converted into stroke parameters interpretable by the robot using the KUKA PRC program. A crucial aspect of this vectorization is ensuring that the raster image consists of clean lines, where curves are represented as long, continuous strokes instead of multiple segments.

To achieve this, we introduced a fine-tuning approach for the Stable Diffusion model, applying context consistency loss during pairwise training to create the Vector LoRA. This Vector LoRA is specifically designed to simplify the image into vector-friendly strokes while maintaining the artistic integrity of the style. A key innovation of our work is the compatibility of Vector LoRA with other Style LoRA modules, enabling us to retain various artistic styles while producing vectorized outputs. This flexibility allowed us to collaborate with renowned Korean cartoon artist Lee Hyun-Se \footnote{Hyun-Se Lee is a first-generation South Korean comic artist best known for his work 'The Terrifying Foreign Baseball Team'. SketcherX was introduced as a robot that draws portraits in the style of Hyun-Se Lee at the exhibition 'The Path of Hyun-Se Lee: The Beginning of the K-Webtoon Legend' in June, 2024}, showcasing his distinctive style through our robot's drawings in an exhibition. Furthermore, the adaptability of Vector LoRA has enabled the system to learn and exhibit a wide range of artistic styles, making it suitable for diverse creative applications.

Capturing the individual characteristics of the subjects was essential to producing personalized portraits. Initially, we attempted to map the captured image into the latent space of a diffusion model via latent inversion, but this approach failed to sufficiently capture individual features or the desired line art style. Therefore, we adopted a couple of complementary approaches. First, we utilized a vision-language model, DeepSeek-VL \citep{lu2024deepseek}, to analyze the image and generate textual descriptions of the subject's age, appearance, accessories, etc., which were then used as input prompts for the diffusion model. Second, we applied ControlNet \citep{zhang2023controlnet} to enhance feature information by extracting Canny edge details from the image, which added morphological edge information to the diffusion process.
By integrating these methods, SketcherX not only creates personalized portraits but also provides an engaging and interactive creative experience, positioning the robot as a novel form of interactive media.

\section{Dynamic Human-Robot Interaction}

SketcherX features two 6-axis KUKA \citep{bischoff2010kuka} robots, one of which serves as the Face Robot. The Face Robot's head is equipped with a camera that detects human faces, initiating image processing upon detection. To create a realistic robotic face, the face content is 3D-modeled using Unity, with a strategically positioned projector behind the face for seamless projection mapping onto the sculpture, aided by display calibration techniques.

User speech is captured by a front-mounted microphone and converted to text via Microsoft Azure's Speech-to-Text (STT) service. The transcribed text is then processed by OpenAI's GPT-4 model \citep{achiam2023gpt}, allowing the robot to respond to user inquiries. During this interaction, the robot analyzes the conversation to detect the user's emotions, triggering corresponding facial expressions—such as laughing, frowning, or crying—and robotic motions like dancing or nodding, enabling dynamic, responsive interactions. 

\section{Method for robotic vector sketching}
\begin{figure*}[t]
\centering
\makebox[\textwidth]{\includegraphics[width=\textwidth]{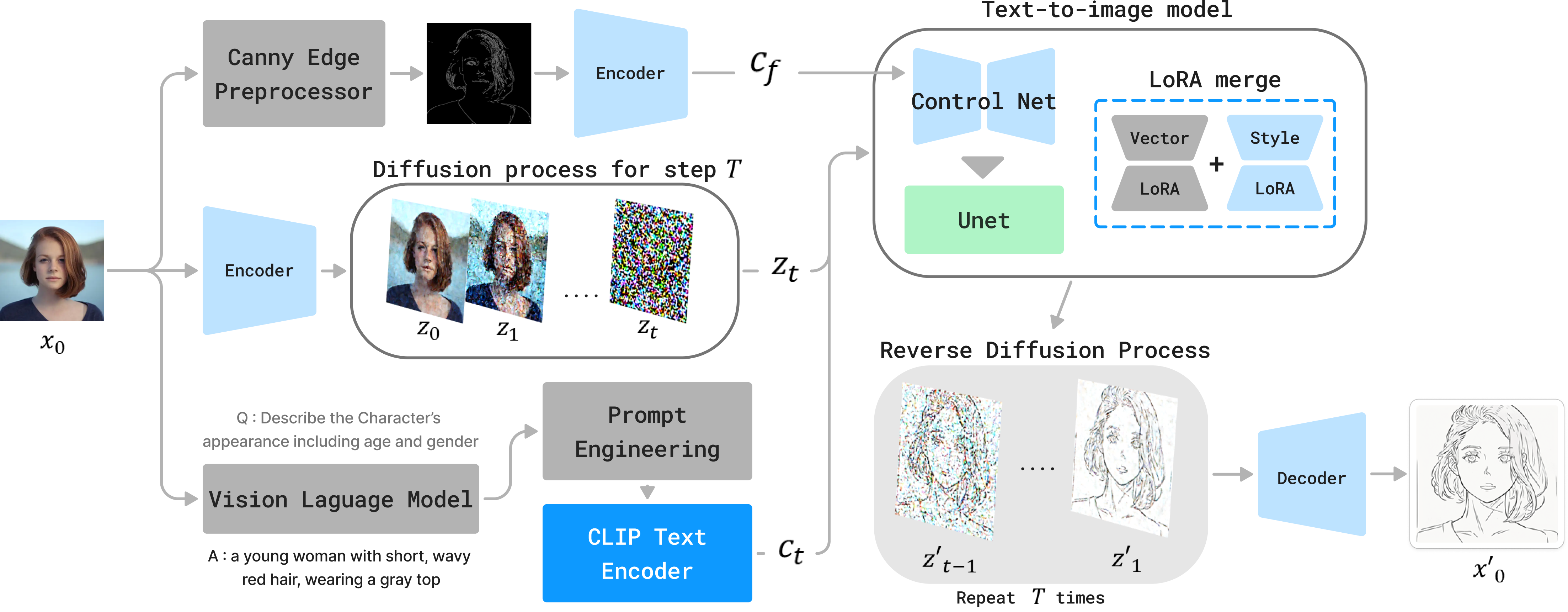}}
\caption{Overview of the robotic vector sketching process. It begins with extracting the user’s facial features using canny edge preprocessor, with its encoded latent vector $c_{f}$fed into ControlNet. Simultaneously, a VL model generates textual descriptions of the user's attributes, which are refined and encoded by the CLIP text encoder into $c_{t}$. The text-to-image model, initialized with latent vector $z_{t}$, is fine-tuned by merging Vector LoRA and Style LoRA, producing vector-friendly strokes while maintaining the desired artistic style.}
\label{fig2}
\end{figure*}

To visually demonstrate the robot's drawing process and create unique portraits, we developed a specialized drawing algorithm. This algorithm had to 1) accurately capture human features in a distinctive drawing style and 2) render the drawing in continuous, sequential strokes. To achieve these preferable conditions, we propose fine-tuning Stable Diffusion models using strategies such as context consistency loss and pair-wise fine-tuning with Dreambooth \citep{ruiz2023dreambooth}. We also integrated ControlNet \citep{zhang2023controlnet} and Vision-language (VL) models to enhance the depiction of human features. The overall drawing process is illustrated in Fig.~\ref{fig2}.

\subsection{Text-to-image model-based stroke line image generation}
To achieve clean and precise vectorization results, we leverage a text-to-image model that outputs well-defined line art, essential for emulating the sequential pen actions of human sketching. This vectorization process is implemented using parameterized basic shapes, such as Bezier curve control points or polygon vertices, making it resolution-independent and allowing for the determination of end-to-end positional points of strokes. We employ the method from \cite{mo2021general}, which utilizes a Recurrent Neural Network (RNN) for step-by-step stroke prediction and a dynamic window to maintain the full resolution of the input. The strokes are generated using a Convolutional Neural Network (CNN) encoder and an RNN decoder, which outputs the stroke parameters. These are then refined through differential rendering based on Bezier curves and differential pasting, converting the resulting SVG files into data interpretable by the KUKA PRC robotic system.

A major challenge is ensuring that the vectorized output is simplified, with curves represented as single, continuous strokes rather than multiple segments. This requires the sketch to maintain consistent line thickness and be free of shadows or shading, ensuring clean lines with precise start and end points.

To achieve this level of precision, we propose fine-tuning the stable diffusion model using our proposed context consistency loss.

\subsubsection{Pair-wise fine-tuning with context consistency loss}
We adopted Dreambooth \citep{ruiz2023dreambooth}, an effective method for fine-tuning with a small number of images, to fine-tune the Stable Diffusion model using a customized LoRA. LoRA \citep{hu2021lora} (Low-Rank Adaptation) is additional trainable low-rank matrices, updating a diffusion model's weight matrix $W$. 
We developed "Vector LoRA," which transforms images into simple stroke line arts suitable for vectorization. This approach allows for easy combination with other style-specific LoRAs, enabling the production of images in diverse styles that are ideal for vectorization.

\begin{figure*}[t]
\centering
\makebox[\textwidth]{\includegraphics[width=1.\textwidth]{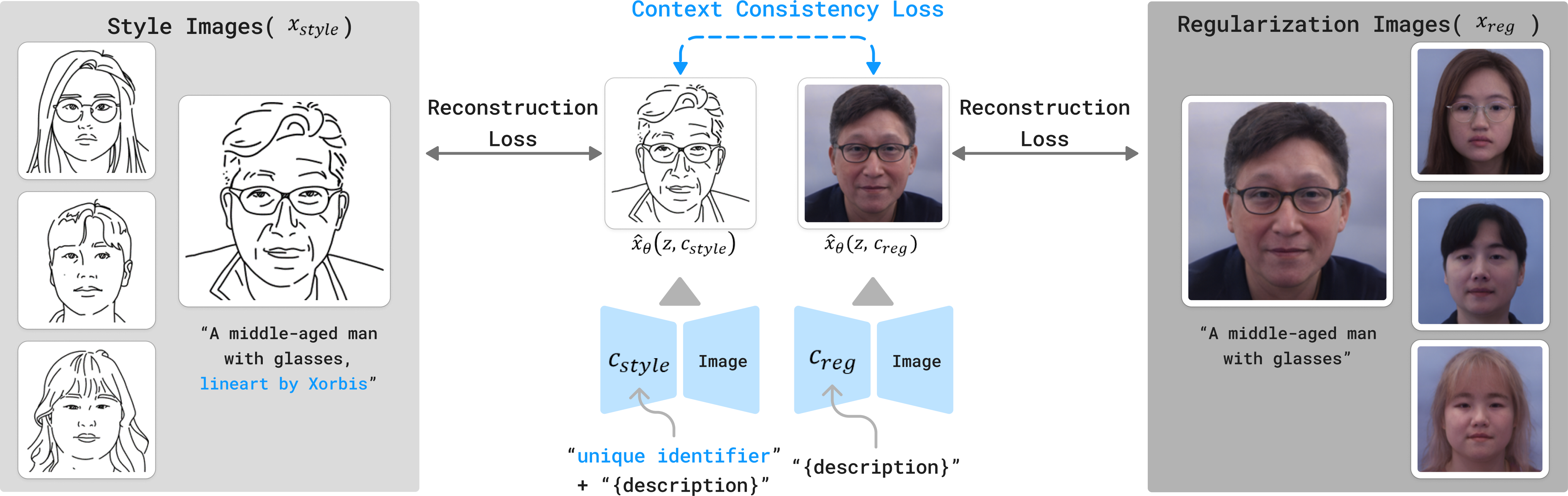}}
\caption{We fine-tuned the model using pair-wise data by applying both context consistency loss and reconstruction loss. For example, if the text prompt for the regularization image $x_{reg}$ is "A middle-aged man with glasses \{description\}" the corresponding text prompt for the style image $x_{style}$ would be "A middle-aged man with glasses \{description\} + lineart by Xorbis \{unique identifier\}," creating a pair-wise data set. Here, $c_{style}$ is the encoded vector of $x_{style}$'s text prompt, and $c_{reg}$ is the encoded vector of $x_{reg}$'s text prompt. The generated image $\hat{x}_{\theta} (z, c_{style})$ is conditioned on $c_{style}$. By calculating the L2 loss between the two generated images, we maintain context consistency during the fine-tuning process.}
\label{figure3}
\end{figure*}

Our findings indicate that pairwise training of regularization images with style images during training enhances the expression of the desired style. In Dreambooth, fine-tuning for a specific style involves appending a unique identifier to the subject's class in the image's text prompt, while applying class-specific prior preservation loss to maintain the semantic prior of the class.

We revised this idea to develop pair-wise training. Our model's input is a face captured by a camera, which we aim to convert into a vectorized image with clear constraints. The training data consists of pairs of style images (lineart drawing of a human face with the desired style) and regularization images (human face data). We denote this pair as ($x_{style}$, $x_{reg}$). The text prompt $P_{style}$ for the style image includes a description and a unique identifier, while the prompt $P_{reg}$ for the regularization image includes only the description. Pre-trained text-to-image diffusion model $\hat{x}_{\theta}$ is fine-tuned using these image pairs, with text prompt pair which is conditioned within a text encoder $\Gamma$, where $c_{style} = \Gamma(P_{style})$ and $c_{reg} = \Gamma(P_{reg})$. Stable diffusion model denoises the latent $z_t = \alpha_{t}x + \sigma_{t}\epsilon$, where $\alpha_{t}$, $\sigma_{t}$ are terminology for noise sampler. We simultaneously trained the model with the ground truth image pair $(x_{style}, x_{reg})$ to enable the model to learn the differences between the two images more closely. The reconstruction loss is as follows: 

\begin{equation}
\begin{split}
    L_{rec} = & \, \lambda_1 \left\|\hat{x}_{\theta}(\alpha_{t} x_{style} + \sigma_{t} \epsilon, c_{style}) - x_{style}\right\|_2^2 \\
    & + \, \lambda_2 \left\|\hat{x}_{\theta}(\alpha_{t} x_{reg} + \sigma_{t} \epsilon, c_{reg}) - x_{reg}\right\|_2^2,
\end{split}
\label{equ_reconstruction}
\end{equation}
where $\lambda_1$ and $\lambda_2$ controls the relative weight for each images in the pair. 

Additionally, we added the context consistency loss, which is the $L_2$ loss of the generated style image and regularization image. In DreamBooth, in order to keep the diversity in image generation and keep the prior knowledge of the subject, it gives the $L_2$ loss $\|\hat{x}_{\theta}(z_t, c_{pr}) - x_{pr}\|^2_2$, where the conditioned vector $c_{pr} = \Gamma(``description")$ and $x_{pr} = \hat{x}_{\theta_0}(z_t, c_{pr})$, which is the generated image from the frozen pre-trained model parametrized by $\theta_0$ conditioned on $c_{pr}$. This is modified to apply $L_2$ loss between pairwise data, and our proposed context consistency loss is defined as follows:

\begin{equation}
    \begin{aligned}
        L_{con} = & \, \left\|\hat{x}_{\theta}(\alpha_{t'} x'_{style} + \sigma_{t'} \epsilon', c_{style}) \right. \\
        & \left. - \hat{x}_\theta(\alpha_{t'} x'_{reg} + \sigma_{t'} \epsilon', c_{reg})\right\|_2^2,
    \end{aligned}
\end{equation}

where $x'_{style} = \hat{x}_{\theta}(z_t, c_{style})$, which is the generated style image conditioned with the text prompt "[description] + [unique identifier]", and $x'_{reg} = \hat{x}_{\theta}(z_t, c_{reg})$, which is the generated regularization image conditioned with the text prompt "[description]". This ensures that the content between the created style image and the reconstructed image is well-preserved,  allowing the model to learn only the differences in drawing style. 
Our overall loss in fine-tuing is as follows:
\begin{align}
    L_{tot} = & \, \mathbb{E}_{x, c, \epsilon, \epsilon', t} \Big[
        w_t \big( \lambda_1 \|\hat{x}_{\theta}(\alpha_{t} x_{style} + \sigma_{t} \epsilon, c_{style}) - x_{style}\|_2^2  \notag \\
        & \quad + \lambda_2 \|\hat{x}_{\theta}(\alpha_{t} x_{reg} + \sigma_{t} \epsilon, c_{reg}) - x_{reg}\|_2^2 \big) \notag \\ 
        & \quad + w_{t'} \|\hat{x}_{\theta}(\alpha_{t'} x'_{style} + \sigma_{t'} \epsilon', c_{style}) \notag \\
        & \quad - \hat{x}_\theta(\alpha_{t'} x'_{reg} + \sigma_{t'} \epsilon', c_{reg})\|_2^2
    \Big].
\label{eq_overall}
\end{align}

For more details on our experiment settings, please refer to the 'Training Dataset and Experiment Settings' in the supplementary section.

\subsubsection{LoRA merging of Vector LoRA and Style LoRA}
\begin{figure*}[t]
\centering
\makebox[\textwidth]{\includegraphics[width=1\textwidth]{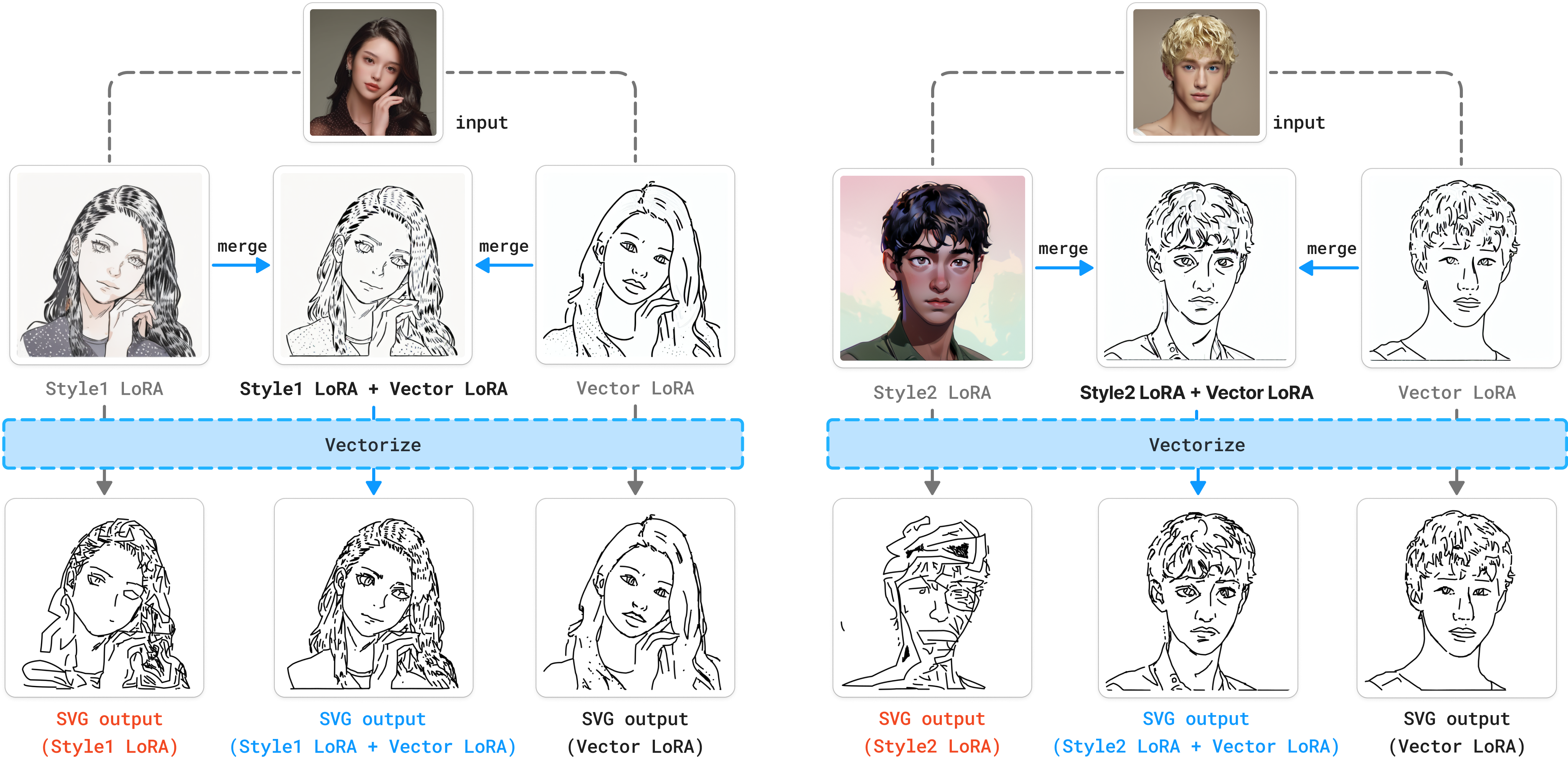}}
\caption{In the figure, Style1 LoRA and Style2 LoRA represent distinct artistic styles. When merged with Vector LoRA, each style is preserved but transformed into clean, vector-friendly strokes. Using only the Style LoRAs results in distorted SVG shapes unsuitable for robotic drawing. However, combining them with Vector LoRA produces clean, continuous strokes, ideal for robotic drawing.}
\label{fig4}
\end{figure*}

Vector LoRA alone can transform an image into a style suitable for vectorization. Moreover, its compatibility with other style-specific LoRAs allows for flexible and diverse stylistic transformations, enabling the creation of images in various styles ideal for vectorization. Specifically, LoRA updates weight matrix $W \in \mathbb{R}^{n \times m}$ in the diffusion model $\hat{x}$ by adding a low-rank term, i.e., $W' = W + BA$, where $B\in\mathbb{R}^{n \times r}$ and $A\in\mathbb{R}^{r \times m}$ for $r < \min(n,m)$. When merged with Vector LoRA and Style LoRA, the updated $W'$ weight of $\hat{x}$ is given by:
\begin{equation}
    \begin{aligned}
        W' &= W + (w_{v} \times B_{v}A_{v} + w_{s} \times B_{s}A_{s}), \\
    \end{aligned}
    \label{eq_lora_merge}
\end{equation}
where $B_{v}$, $A_{v}$ are the matrices of low-rank factor in Vector LoRA, $B_{s}$, $A_{s}$ are the matrics of Style LoRA and $w_v$ and $w_s$ are balancing parameters. Fig.~\ref{fig4} compares the SVG conversion results when using only the Style LoRA and when using both the Vector LoRA and Style LoRA. The results demonstrate that Vector LoRA is highly effective for generating smooth stroke lines. Additionally, it shows that Vector LoRA is compatible with various Style LoRAs, allowing for diverse stylistic transformations. 
This flexibility allowed us to collaborate with renowned manga artist Lee Hyun-Se, creating a robot exhibition that drew portraits in his distinctive style. The exhibition received significant attention and spotlight from various media outlets.  

\subsection{Techniques for Human Feature Extraction}
Accurately capturing human features is essential for portrait drawing. Initially, we use StyleGAN v2 \citep{karras2020analyzing} with StyleCLIP \citep{patashnik2021styleclip} to modify features such as expressions and accessories. However, these models often overlook details not present in the CelebA-HQ dataset, such as clothing details and diverse accessories, resulting in limited diversity. Furthermore, the requirement for precise facial alignment makes it challenging to capture upper body features.

As an alternative, we adopt a diffusion-based approach because of its ability to capture diverse features accurately. Among the methods for transforming real images, Stochastic Differential Editing (SDEdit) \citep{meng2021sdedit} is most commonly used. It adds noise to the input image and iteratively denoises it through a stochastic differential equation. However, SDEdit alone cannot fully preserve the source image's characteristics and achieve the desired style transfer. To address this, we added ControlNet to enhance the representation of human features by incorporating canny edge information as a condition, resulting in more accurate portraits. Additionally, we utilize Vision-Language model to extract and tag distinguishing features such as age and appearance, which are then reflected to the text prompt using prompt engineering. Fig.~\ref{fig2} shows the examples of query. For detailed information about the model, please refer to the supplementary section.

\section{Conclusion}
We developed an innovative robotic system for creating personalized portraits through interactive user engagement. Combining advanced diffusion-based image processing with real-time human-robot interaction, our platform captures a person's likeness with a head-mounted camera, processes the image into a distinctive artistic style, and uses a robotic arm to draw the portrait in a human-like manner.

Unlike previous robotic art methods that rely on analog printing, our system emphasizes the robot's autonomous identity and interactivity. Integrating Large Language Model (LLM) technology, the robot can engage in meaningful conversations, enhancing user experience. The custom drawing algorithm, fine-tuned stable diffusion model, and ControlNet and Vision Language models ensure portraits are unique and true to the subject's features.

Optimized for efficiency, our system delivers high-quality portraits in about two minutes, with the Vector LoRA enabling seamless adaptation to various artistic styles. This work showcases a new paradigm in robotic creativity, where robots actively participate in the artistic process, moving beyond mere reproduction. Future work will further refine the system's artistic capabilities and explore broader applications in creative domains.

{
    \small
    \bibliographystyle{ieee_fullname}
    \bibliography{references}
}

\clearpage

\section{Supplementary Materials}
\subsection{Hardware composition}
\begin{figure*}[t] 
    \centering
    \includegraphics[width=0.95\textwidth]{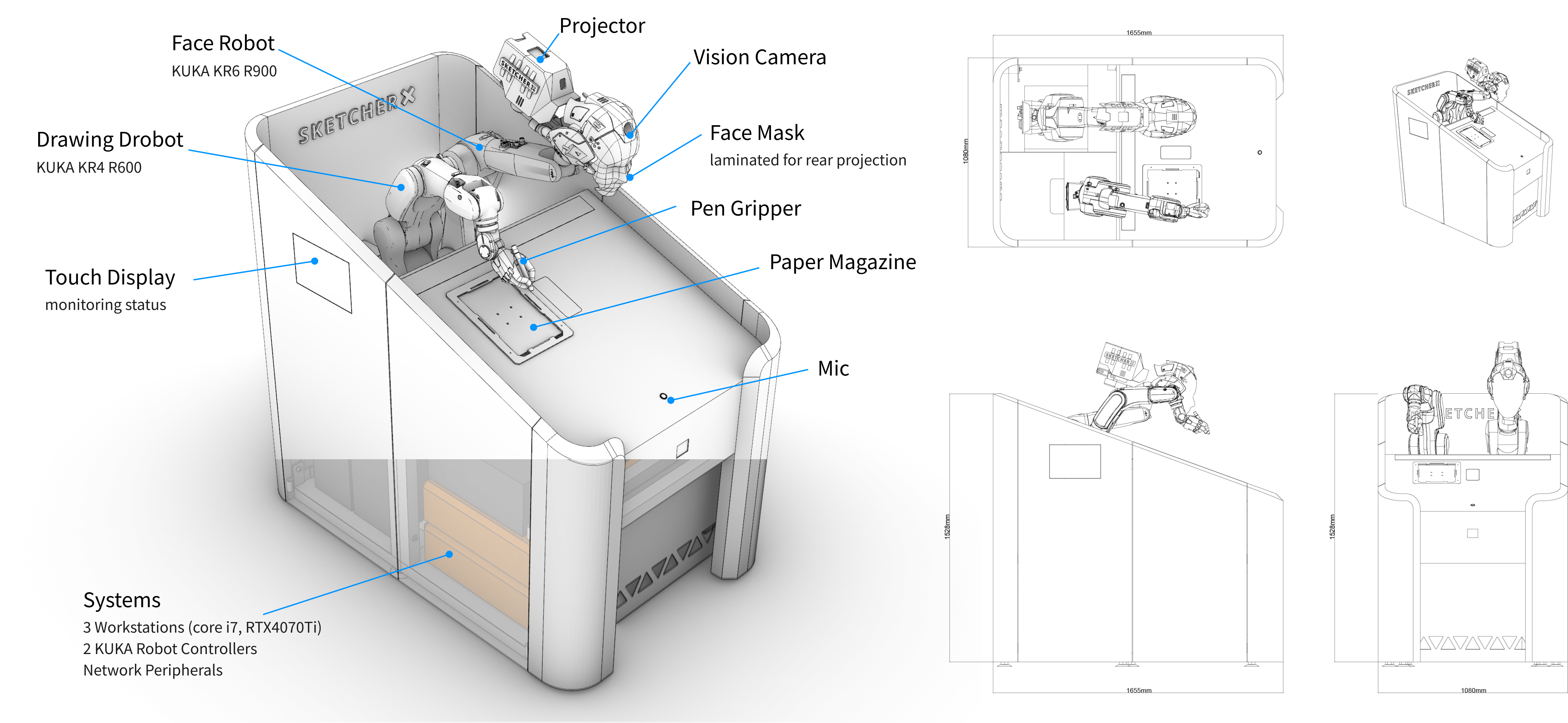} 
    \caption{Hardware configuration and system schematic illustrating the setup of SketcherX.}
    \label{sup_fig3}
\end{figure*}
The hardware configuration for the robot system is as follows. The detailed hardware composition is depicted in Figure \ref{sup_fig3}. We utilized two KUKA (KR4, KR6) models to create a Drawing Robot and a Face Robot. The Face Robot is equipped with a Face Mask at its end effector, where a 3D face is mapped onto the mask using the Unity engine, facilitated by a projector positioned behind the mask. An RGB camera is mounted on the top of the mask, simulating the robot's head, to capture human facial images.

The Drawing Robot is equipped with a pen gripper, fabricated using a 3D printer, at its end effector. A pen is inserted into this gripper, enabling the robot to draw. The system is supported by three workstations, each equipped with Core i7 processors and RTX 4070Ti GPUs. One workstation handles machine learning processing, another is responsible for face mapping, and the third serves as the master PC, managing communication between the KUKA robots and the workstations, as well as overseeing the entire system. DeviceNet is used to transmit coordinate data from the PC to the robot control system, ensuring precise execution of tasks.

\subsection{Training Dataset and Experiment Settings}
\begin{figure*}[t] 
    \centering
    \includegraphics[width=0.95\textwidth]{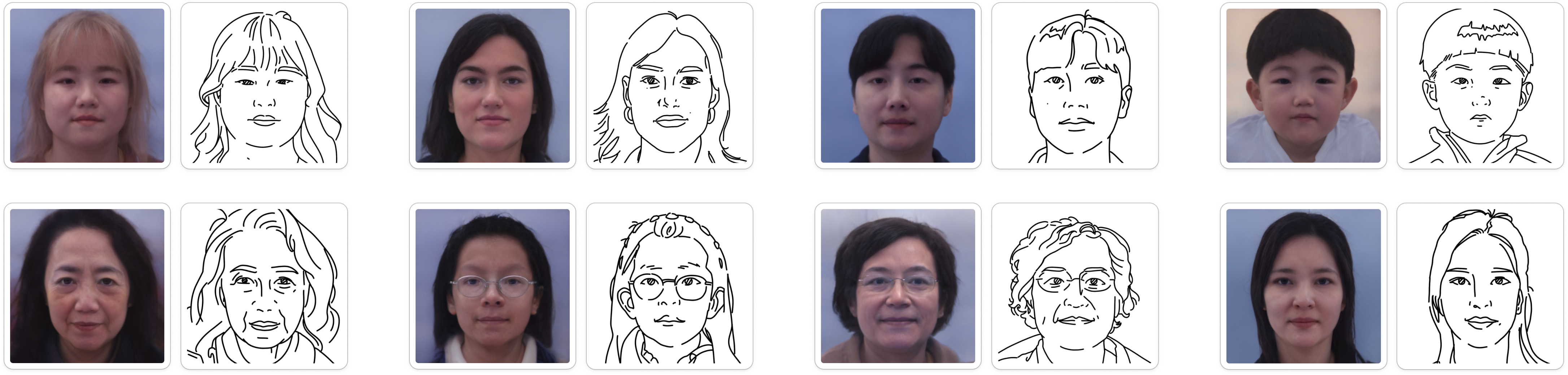} 
    \caption{Examples of pair-wise dataset for training Vector LoRA.}
    \label{sup_fig5}
\end{figure*}
To create the Vector LoRA, we employed the DreamBooth fine-tuning method. Initially, we prepared a dataset consisting of 50 image pairs. After training the model, we further refined it using augmented data generated by the model itself. Each image pair consists of a regularization image, where a person is facing forward, and a style image, which is a traced version of the same forward-facing photograph. The tracing technique was used for the style image to ensure that the vectorized image is composed of a continuous one-line stroke outlining the person. For the regularization images, we generated virtual individuals facing forward using StyleGAN v2. Figure \ref{sup_fig5} illustrates an example of these image pairs.

During the first 500 iterations, we trained the model using the initial 50 image pairs. For the remaining 500 iterations, we incorporated an additional 200 augmented data pairs generated by the model. The weight terms for the reconstruction loss \ref{equ_reconstruction}, denoted as $\lambda_{1}$ and $\lambda_{1}$, were both set to 0.5. The weights for the overall loss \ref{eq_overall}, the $w_{t}$ of the reconstruction loss and $w_{t'}$ of the context consistency loss were also set to 0.5 during the first 500 iterations. However, for the subsequent 500 iterations, we adjusted the weights to 0.8 for the reconstruction loss and 0.2 for the context consistency loss, focusing on enhancing the precision of the output.

\subsection{Performance Comparison Across Different Artistic Styles}
\begin{figure*}[t]
    \centering
    \includegraphics[width=0.95\textwidth]{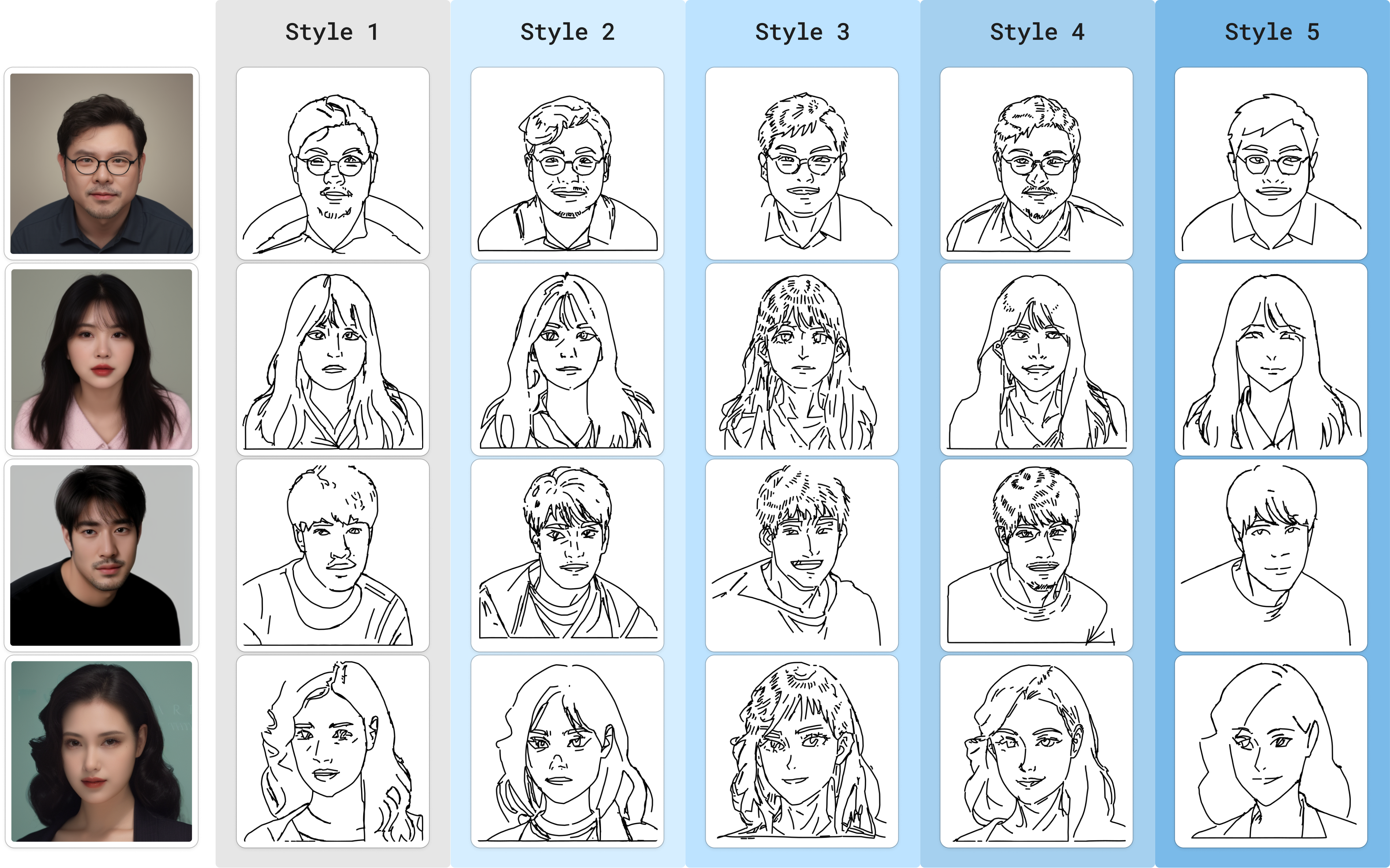}
    \caption{SVG output of five different artistic styles. Each column indicates different model inference outputs, which the model's weight was updated with different Style LoRA + Vector LoRA.}
    \label{sup_fig1}
\end{figure*}
The Figure \ref{sup_fig1} presented is an SVG output created by combining five different Style LoRAs with a Vector LoRA. Each column represents outputs with the same style. As discussed in the main text, our stroke transformation is crucial for producing clean lines composed of a single stroke, ensuring that the robot can accurately render the image. The role of the Vector LoRA is to facilitate vectorization while maintaining the distinctive style of each Style LoRA. When the figure is enlarged, the clarity and precision of the lines that compose each image become evident. Since the figure is the result of vectorization, it accurately corresponds to what the robot would draw on paper. The main contribution lies in the ability to effectively capture human features while transforming them into the desired style.

\subsection{Impact of ControlNet}
\begin{figure*}[b] 
    \centering
    \includegraphics[width=0.95\textwidth]{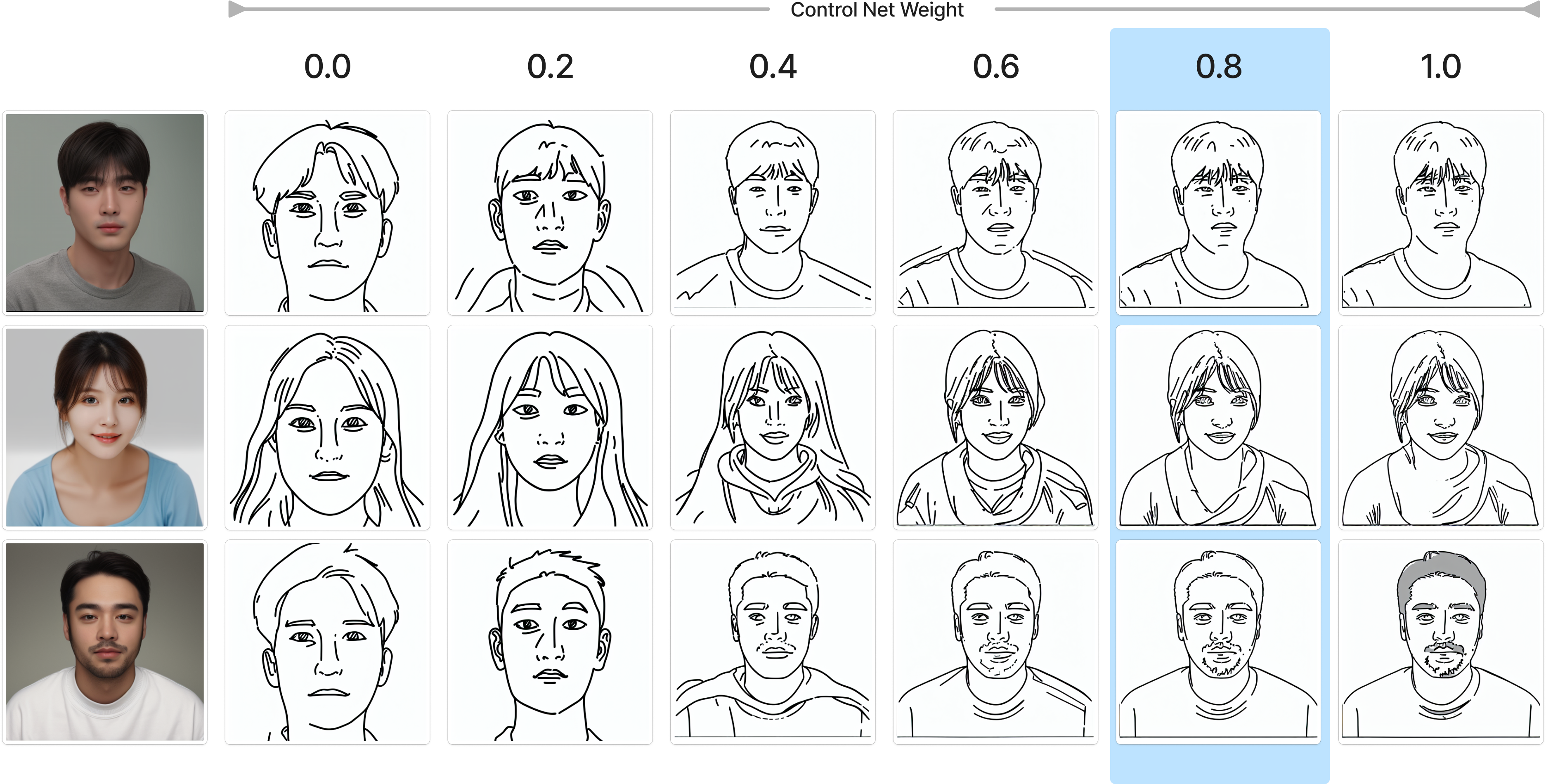}
    \caption{Comparison image output depending on different ControlNet weight. The drawing output was generated using only Vector LoRA, without any additional Style LoRA.}
    \label{sup_fig2}
\end{figure*}
Using the SDEdit method, specifically the img2img approach alone, was insufficient to capture all the unique characteristics of the user in a precise portrait. Figure \ref{sup_fig2} compares the different outcomes achieved by adjusting the ControlNet weight. When the weight is set to 0, the portrait reflects only coarse characteristics, such as the user's gender, but fails to capture deeper features. We found that a weight of 0.8 produced the most optimal results. However, this weight may need to be adjusted depending on the style LoRA used in conjunction with it.

\subsection{Artwork Examples and Exhibition History}
\begin{figure*}[t]
    \centering
    \includegraphics[width=0.95\textwidth]{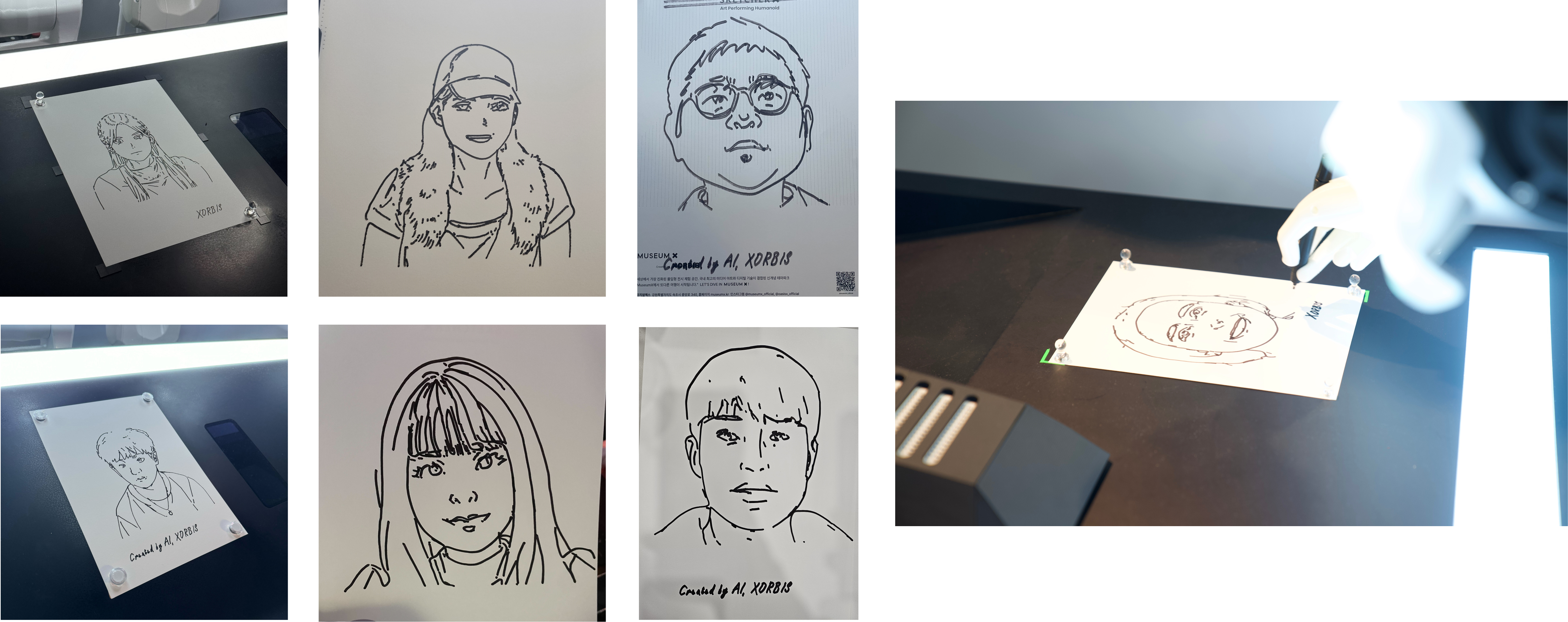}
    \caption{Examples of actual portraits created by SketcherX.}
    \label{sup_fig6}
\end{figure*}

\begin{figure*}[t]
    \centering
    \includegraphics[width=0.95\textwidth]{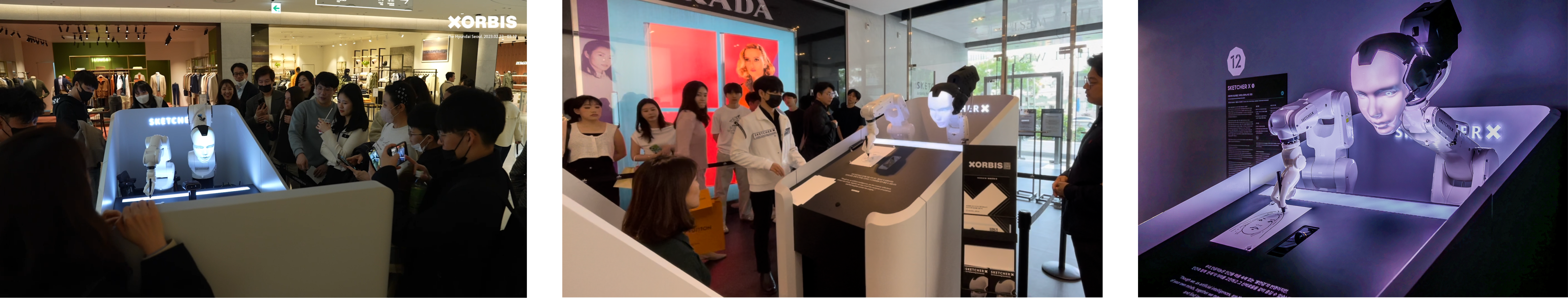}
    \caption{A depiction of SketcherX during an actual exhibition, showcasing its interaction with attendees.}
    \label{sup_fig7}
\end{figure*}

Figure \ref{sup_fig6} shows the examples of portraits created by SketcherX. These portraits effectively capture the unique characteristics of individuals, adapting to various styles based on the nature of the exhibition. SketcherX tailors its artistic approach to suit the theme of each exhibition, producing portraits that resonate with the audience.

Figure \ref{sup_fig7} is a depiction of SketcherX during an actual exhibition, showcasing its interaction with attendees. Since its debut at CES 2023, SketcherX has undergone continuous development, evolving into its current form. It has been exhibited at various venues, including The Hyundai Seoul, Galleria Department Store in Apgujeong, Hyundai Department Store in Pangyo, Lee Hyun-se's Road: The Legend of K-Webtoon, and Museum-X in Sokcho. Through these exhibitions, SketcherX has engaged with the public and received positive recognition for its approach as an interactive media.
\section{Credit Attributions}

This project was performed by XORBIS R$\&$D Center. 
We would like to express our sincere gratitude to the following individuals for their invaluable contributions to this project. Sukhwan Choi for system integration and robotics programming, Jeewoong Lieu for product and mechanical design, Jiwoong Ryu and Kimyung Lee for face software and 3d programming, Solmi Kim and Gyeongwon Joo for robot face and 3d modeling design, and Seunghun Mok for chat system integration. We also acknowledge Eunjung Yoo and Chaewon Kim for curation, Dr. Kyounghun Lim for design, Dr. Jungwoo Chae for his support in creative directing. 







\end{document}